\documentclass[11pt,a4paper]{article}

\usepackage[hyperref]{acl2020}
\usepackage{times}
\usepackage{latexsym}

\usepackage{etoolbox}

\usepackage[utf8]{inputenc} 
\usepackage[T1]{fontenc}
\usepackage[vietnamese]{babel} 
\usepackage{CJKutf8}
\usepackage{booktabs}
\usepackage{natbib}
\usepackage{graphicx}
\usepackage{amsmath}
\usepackage{amssymb}

\def\ngramlstm{n\mathrm{\_gramLSTM}} 
\def\nigramlstm{n\mathrm{\_gramLSTM}} 
 
\def\nigram{m_t\mathrm{\_gram}} 
\DeclareMathOperator{\zip}{zip} 
\DeclareMathOperator{\unzip}{unzip} 
\DeclareMathOperator{\Phrase}{Phrase}

\usepackage{microtype}

\aclfinalcopy % Uncomment this line for the final submission
\title{An Effective Method using Phrase Mechanism in \\Neural Machine Translation}

 \author{ Phuong Nguyen \and  Minh Le Nguyen\\
 Japan Advanced Institute of Science and Technology, Japan  \\
\texttt{\{phuongnm, nguyenml\}@jaist.ac.jp} \\}

\date{}

\begin{document}
\maketitle
\begin{abstract}
    Machine Translation is one of the essential tasks in Natural Language Processing (NLP), which has massive applications in real life as well as contributing to other tasks in the NLP research community.
    Recently,  Transformer -based methods have attracted numerous researchers in this domain and achieved state-of-the-art results in most of the pair languages. In this paper, we report an effective method using a phrase mechanism, PhraseTransformer,  to improve the strong baseline model Transformer, in constructing a Neural Machine Translation (NMT) system for parallel corpora Vietnamese-Chinese. Our experiments on the MT dataset of the VLSP 2022 competition achieved the BLEU score of 35.3 on Vietnamese to Chinese and 33.2 BLEU scores on Chinese to Vietnamese data. 
    Our code is available at \url{https://github.com/phuongnm94/PhraseTransformer}.
% 
%  anonymous  

\end{abstract}

\section{Introduction}

In the NLP area, Machine Translation is the primary task that has a long time history of development, especially in the approaches using Neural Networks \citep{NIPS2014_a14ac55a,jointattn}. Besides, there are numerous proposed architectures in this domain that have got a significant effect on the NLP community in many domains \citep{NIPS2014_a14ac55a,jointattn,transformer,devlin-etal-2019-bert}.  Therefore, in this work, we
% anonymous - team JAIST NLP (JNLP) 
focus on Machine Translation task in \textit{The Ninth International Workshop on Vietnamese Language and Speech Processing (VLSP 2022)\footnote{\url{https://vlsp.org.vn/vlsp2022}}.}

VLSP 2022 is a competition of one of the biggest NLP communities in Vietnam, \textit{Association for Vietnamese Language and Speech Processing}, hosted by the VNU University of Science, Hanoi.
 This is a good chance for researchers in many domains to introduce their awesome research results and catch the state-of-the-art machine learning models in different domains.   
  
This year, the machine translation task is designed to translate from Vietnamese to Chinese and on the contrary direction. In this task, given the input is a natural sentence in Chinese (or Vietnamese), the machine translation system is required to generate a new sentence in Vietnamese (or Chinese) that has the same meaning as the input sentence (Table~\ref{tab:example}). 
 
\begin{CJK*}{UTF8}{gbsn} 
\begin{table}[ht]
\small
\centering
\begin{tabular}{lp{0.34\textwidth}}
    \toprule
     \textbf{Language}  & \multicolumn{1}{c}{\textbf{Content}}   \\   \midrule
    Vietnamese     &    Tôi vừa có kế hoạch thay họ biểu diễn ở bữa tiệc. \\ 
    Chinese & 正打算让这群废物在派对上表演呢 \\  
    \textit{(meaning)} & \textit{I just had the plan to perform on their behalf at the party.}\\\midrule
    Vietnamese &  Không! Cậu đã lấy đi tất cả những gì quý giá đối với tôi rời, cậu Toretto à! \\ 
    Chinese & 我有价值的东西早被你榨取光了 托雷托先生 \\   
    \textit{(meaning)} & \textit{No! You took everything that was precious to me, Mr. Toretto!} \\
    \bottomrule
\end{tabular}
\caption{Example of machine translation task in the parallel Vietnamese-Chinese corpus of VLSP 2022. The \textit{(meaning)} row is to support the readers of this paper which are not provided in dataset.}
\label{tab:example}
\end{table} 
\end{CJK*}

Recently, the self-attention mechanism has occupied a lot of attention, especially the Transformer architecture \citep{transformer,devlin-etal-2019-bert}. This architecture leads many state-of-the-art results in various NLP domains  \citep{devlin-etal-2019-bert,lewis-etal-2020-bart,JMLR:v21:20-074}.  In this work, we applied a new approach based on the Transformer architecture, PhraseTransformer \citep{Nguyen2023}, which leverages a phrase attention mechanism to solve the machine translation task. The main idea of this approach is to enhance the word representation by its local contexts (or phrases), and apply the self-attention mechanism to model the dependencies between phrases in a sentences. The experimental results on the machine translation dataset of the VLSP 2022 workshop show that our PhraseTransformer clearly beat the original Transformer model in both two directions from Vietnamese to Chinese and from Chinese to Vietnamese. Furthermore, the PhraseTransformer does not require any external syntax tree information as the previous works \citep{yang-etal-2018-modeli,wang-etal-2019-tree,TreeStructured,bugliarello-okazaki-2020-enhancing} and is more lightweight compared with other phrase-level attention models \citep{xu-etal-2020-learning}. To this end, our PhraseTransformer improved the original Transformer by 1.1 BLEU scores on the setting from Vietnamese to Chinese and 1.3 BLEU scores on the setting from Chinese to Vietnamese.

The remainder of this paper, we detail our work in four sections.  Section~\ref{sec:relatedW}, we show the related works in this domain. And describe the details of our PhraseTransformer architecture  in section~\ref{sec:model_architecture}. Then, the experiments and result analysis are shown in section~\ref{sec:exp}. Finally, in section~\ref{sec:conclusion}, we conclude this paper.

\section{Related works \label{sec:relatedW}}

In the success of Transformer \cite{transformer} in the machine translation task, there are a numerous works proposed new directions to improve this architecture. Among them, the approach considering of utilize the phrase linguistic structure representation is promising. The works presented in~\citet{yang-etal-2020-improving,Shang2021GuidingNM,10.1007/978-3-031-08530-7_25} demonstrate that the template prediction can guide an NMT system in the decoding process to improve performance. The works  \citep{wu-etal-2018-phrase,TreeStructured,wang-etal-2019-tree} indicate that the phrase information extracted from the syntax tree can improve the meaning representation of sentence. However, the performance of these systems is affected by the quality of the syntax tree extraction step which is usually lower performance in low-resource languages.  Close to our PhraseTransformer, \citet{xu-etal-2020-learning} also presents a different method to model the phrase in a source sentence, however, this architecture uses a huge number of parameters to learn the attention scores between source phrases with target words. Compared with this work,  by using LSTM architecture \citep{lstm_hochreiter1997long} in the Multi-head layers to model various local context information, our PhraseTransformer increases the model size with a small margin but works effectively.  
\section{Model architecture \label{sec:model_architecture}}
\begin{figure*}[!htbp]
            \centering 
            \centerline{\includegraphics[width=0.48\textwidth, keepaspectratio,
            trim={0cm 0 0 0.0}, clip=true]{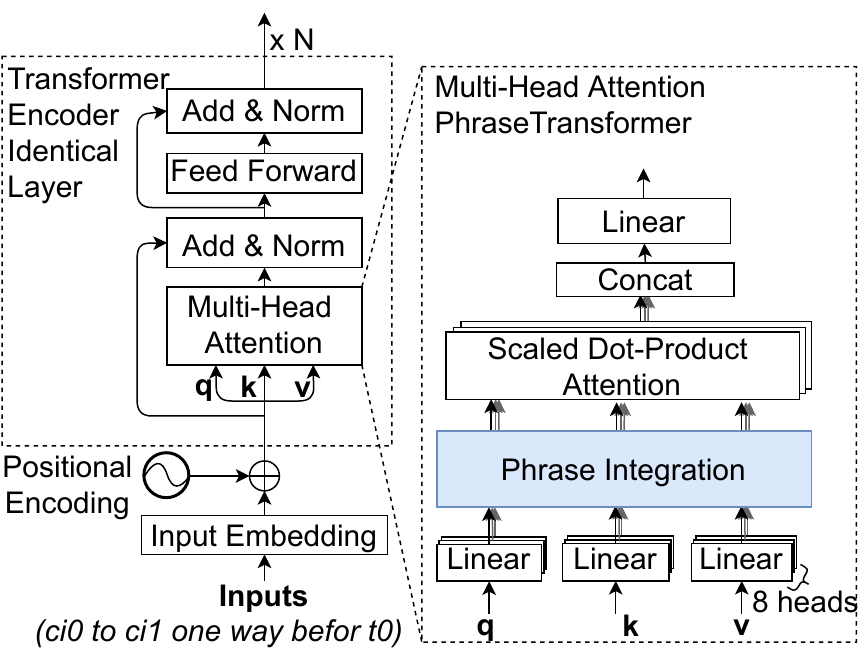}   
            \includegraphics[width=0.48\textwidth, keepaspectratio,
            trim={0cm 0 0 0.0}, clip=true]{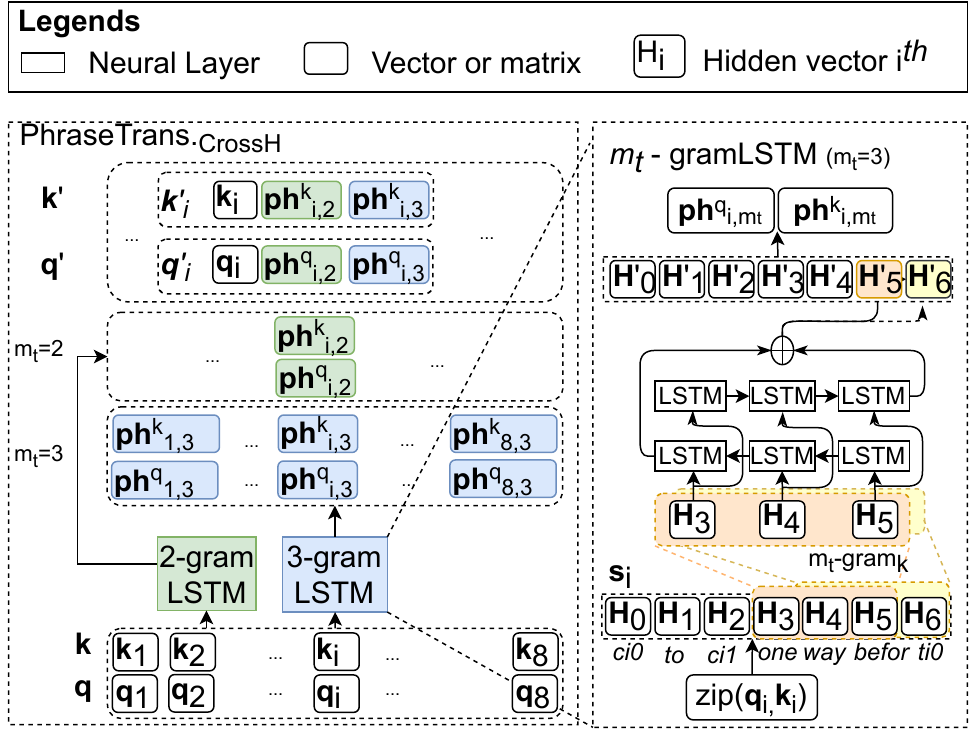}}
            \caption{Overview of PhraseTransformer (CrossH) using $n$\_gram LSTM in  MultiHead layer. In this case, the phrase representations are built  with $\mathrm{gram\_size} = \{2,3 \}$, 2-gram, 3-gram  models apply to all 8 heads. 
            }
            \label{fig:overviewPhrasetrans}
        \end{figure*}

    In this section, we describe our PhraseTransformer model \citep{Nguyen2023}, which is used to submit the machine translation task in VLSP 2022 workshop (Figure~\ref{fig:overviewPhrasetrans}). We define the input vector ($\mathbf{x}$) is synthesized from the word embedding and positional encoding  $\mathbf{x} = [\mathbf{x}_1, ... , \mathbf{x}_{\vert S \vert}] $   ($\vert S \vert$ is sentence length) similar to \citet{transformer}. This vector is processed by a Dense layer to get multi-views of input data corresponding to different heads. 
    \begin{align} 
            \mathbf{q}_i, \mathbf{k}_i, \mathbf{v}_i &= \mathbf{x}\mathbf{W}_i^q , \mathbf{x}\mathbf{W}_i^k , \mathbf{x}\mathbf{W}_i^v  \label{eq:qkv_tf}
    \end{align}
    where $i$ is the    head index, $   1 \leq i \leq{H}$ where $H$ is the number of heads in Multi-Head layer. We define a $\Phrase$ function to extract local context:   
        \begin{align} 
        \resizebox{0.88\columnwidth}{!}{%
            $
            \Phrase(\mathbf{s}_i, m_t) =  
                \begin{cases}
                    \ngramlstm(\mathbf{s}_i) &\mathrm{if\quad} n_{i} > 0 \\
                    \mathbf{s}_i  &\mathrm{otherwise}
                \end{cases} \label{eq:phrase_define} $} 
        \end{align}
        where  $m_t$ is the hyper-parameter gram size,  $\mathbf{s}_i$ is the sequence of words vector,  $\nigramlstm_k$ is the function which capture local context using LSTM model:
        \begin{align}
            \nigram_k(\mathbf{s}_i) &= [\mathbf{H}_{k-n_i+1},\mathbf{H}_{k-n_i+2}, ...,\mathbf{H}_{k}] \label{eq:ngramk} \notag\\
            \nigramlstm_k(\mathbf{s}_i)  &= \mathrm{LSTM}^f_i(\nigram_k(\mathbf{s}_i)) +  \notag  \\
            &\quad\quad \mathrm{LSTM}^b_i(\nigram_k(\mathbf{s}_i)) \notag
        \end{align}
       Then, the phrase information is integrated to the word representations: 
        \begin{align}
            \zip (A, B) &= [ A^{\intercal};B^{\intercal} ]^{\intercal} \\
            \mathbf{ph}^k_{i, m_t}, \mathbf{ph}^q_{i, m_t}  &= \unzip( \Phrase(\zip(\mathbf{q}_i, \mathbf{k}_i),  m_t))\label{eq:zipqk} \\
            \mathbf{ph}^q_{i} &= \zip(\{\mathbf{ph}^q_{i, m_t} \mid m_t \in \mathbf{m}  \})\\
            \mathbf{ph}^k_{i} &= \zip(\{\mathbf{ph}^k_{i, m_t} \mid m_t \in \mathbf{m}  \})
        \end{align}
         After that, these vectors are fed to the Self-Attention layer to learn the dependencies between words and phrases.
        \begin{align} 
             \mathbf{head}_i &= \mathrm{softmax}( \frac{\mathbf{ph}^q_{i}\,\, {\mathbf{ph}^k_{i}}^\intercal }{\sqrt{d_h}})\,\mathbf{v}_i
        \end{align}
        where $d_h$ is  dimension of one head in Multi-Head layer. Finally, all head vector representation is combined to get the final sentence vector ($\mathbf{h}_{out}$) following \citet{transformer}: 
        \begin{align}
            \mathbf{h}_{mh} &= [\mathbf{head}_1; ...; \mathbf{head}_H] \mathbf{W}^o \\
            \mathbf{h}_{no} &=\mathrm{LayerNorm}(\mathbf{h}_{mh} + \mathbf{x}) \label{eq:Addnorm} \\ 
            \mathbf{h}_{out} &=\mathrm{LayerNorm}(\mathrm{FeedForward}(\mathbf{h}_{no}) + \mathbf{h}_{no})
        \end{align}
        This sentence vector is processed via a stack of $N$ Identical layers with the same architecture and forwarded to the Transformer Decoder layer \citep{transformer} for decoding process.
        
        \paragraph{\textbf{Training method}} The training objective is to maximize the Log-Likelihood function of the probabilities to  generate the target sentence ($y$) given a source sentence ($x$) from an machine translation parallel dataset ($\mathcal{D}$): 
         \begin{align}
            \mathrm{maximize}: \sum_{<x, y> \in \mathcal{D}} \log\, p_{\theta}\, (y \mid x) 
        \end{align}
\section{Experiments and Results \label{sec:exp}}
    In this section, we describe in detail about our experiments and the corresponding results.
    \paragraph{Dataset.} We use the parallel corpora Chinese-Vietnamese provided by the VLSP 2022 workshop for the training and development process. For more understanding of this dataset, we show the data analysis information on Table~\ref{tab:data_detail}.  
        \begin{table}[htpb]
            \centering 
            \resizebox{\columnwidth}{!}{%
            \begin{tabular}{cccccccc}
                \toprule 
                  \multicolumn{3}{c}{\textbf{\#examples}}&\multicolumn{2}{c}{\textbf{\#vocab}}&\multicolumn{2}{c}{\textbf{Avg. length}} \\
                 Train & Dev & Test & Vi & Zh & Vi & Zh\\\midrule  
                 300,347 & 999 &  999 &4K & 16K & 18.99 &  22.85    \\   \bottomrule
            \end{tabular}
            }
            \caption{
            Statistics information of parallel corpora Chinese-Vietnamese in VLSP 2022 workshop. Vocabulary size and average length of source (Src) and target (Tgt) side are computed on training set.  \label{tab:data_detail}}  
        \end{table} 
    \paragraph{Prepossessing.} We follow the previous work \citep{provilkov-etal-2020-bpe} for reprocessing step. To deal with the out-of-vocab problem, we use BPE encode method \citep{bpe} to split words into sub-tokens with the number of operators is 4000 and 16000 operators  in Vietnamese (Vi) and Chinese (Zh) languages, respectively. In the Chinese language, because there is no space between characters, the BPE segment module directly processes the whole raw sentence as a word segment. 
    
    \paragraph{Evaluation metric.} We use the BLEU score to evaluate the quality of the translated sentence.  For a fair comparison with other works, we used SacreBleu \citep{post-2018-call} tool\footnote{\url{https://github.com/mjpost/sacrebleu}}. To evaluate the system translated to the Chinese language, we used a word tokenizer default provided by this tool with setting \texttt{--tok zh}. To this end, the output of our machine translation system is compared with the raw sentence of the target language provided by VLSP 2022 parallel corpora to get the BLEU score.  The signature of our evaluation setting on both translation directions are: 
    \begin{itemize}
        \item Vietnamese-Chinese: \texttt{BLEU+case.mixed +lang.vi-zh +numrefs.1 +smooth.exp +tok.zh +version.1.5.1}
        \item Chinese-Vietnamese: \texttt{BLEU+case.mixed +lang.zh-vi +numrefs.1 +smooth.exp +tok.13a +version.1.5.1}
    \end{itemize}
    
    \paragraph{Experimental setting.} For comparison between our PhraseTransformer and the original Transformer \cite{transformer}, we implemented to train and evaluate these models on the parallel corpora Chinese-Vietnamese VLSP 2022 without any external data or pre-trained model. We conducted all experiments with the same setting on server NVIDIA A40. The hidden size is $512$; the learning rate is initialized by $7e^{-4}$; the learning warm-up step is $6000$; both the transformer encoder and decoder contain 6 layers, 4 heads. In the training process, the batch size is 4096 tokens, and the maximum number of epochs is $100$. Following \citet{provilkov-etal-2020-bpe}, we average 5 latest checkpoints\footnote{The script is used to get average 5 latest checkpoints: \url{https://github.com/facebookresearch/fairseq/blob/main/scripts/average_checkpoints.py}} to get a final model for the translation testing process.
    
    \paragraph{Results.}  
        We report the the experimental results of machine translation systems from Chinese to Vietnamese and from Vietnamese to Chinese on Table~\ref{tab:result1}, and Table~\ref{tab:result2}, respectively\footnote{The underline setting in these tables is the model we submitted at the time of the VLSP 2022 competition.}. 
        Based on these results, we found that our PhraseTransformer can beat the original Transformer with almost settings of gram sizes. These results proved the effectiveness of our phrase modeling mechanism, which supports the translation system better in sentence meaning representation. Besides, our model is work well without any external syntax tree information that makes the PhraseTransformer architecture can widely adaptable to many other languages as well as other NLP tasks.
        \begin{table}[htpb]
            \centering 
            \resizebox{\columnwidth}{!}{%
            \begin{tabular}{llcc}
                \toprule 
                  \multicolumn{1}{c}{\textbf{Model}}&\textbf{gram sizes ($\mathbf{m}$)}&\multicolumn{1}{c}{\textbf{Dev}}&\multicolumn{1}{c}{\textbf{Test}} \\\midrule
                %  Train & Dev & Test & Vi & Zh & Vi & Zh\\\midrule  
                Transformer & - &  30.4  & 31.9 \\ 
                PhraseTransformer & \{3\} & \textbf{31.0}  & 33.0 \\ 
                PhraseTransformer & \{4\} &  30.9  & 32.7 \\ 
                \underline{PhraseTransformer} &  \{2,3\} &  30.8  & \textbf{33.2} \\ 
                \bottomrule
            \end{tabular}
            }
            \caption{Experimental result using SacreBLEU of translation system from Chinese to Vietnamese. The test set is the public test provided at the time of the competition.\label{tab:result1}}  
        \end{table}  
        \begin{table}[htpb]
            \centering 
            \resizebox{\columnwidth}{!}{%
            \begin{tabular}{llcc}
                \toprule 
                  \multicolumn{1}{c}{\textbf{Model}}&\textbf{gram sizes ($\mathbf{m}$)}&\multicolumn{1}{c}{\textbf{Dev}}&\multicolumn{1}{c}{\textbf{Test}} \\\midrule
                %  Train & Dev & Test & Vi & Zh & Vi & Zh\\\midrule  
                Transformer & - &  29.5  & 34.2 \\ 
                PhraseTransformer & \{3\} & \textbf{30.1}  & 35.0 \\ 
                \underline{PhraseTransformer} &  \{4\} &  \textbf{30.1} & \textbf{35.3} \\ 
                PhraseTransformer & \{2,3\} &30.0  & 34.1  \\ 
                \bottomrule
            \end{tabular}
            }
            \caption{The experimental result using SacreBLEU of translation system from Vietnamese to Chinese. 
            % The test set is the public test provided at the time of the competition.
            \label{tab:result2}}  
        \end{table}    
    \paragraph{Improvement example.}
    For more understanding about the improvement of PhraseTransformer compared with the original Transformer, we observe the difference between the outputs of two models and show it in Table~\ref{tab:improvement}. In the first row, the original Transformer has missed a part of the information in the translation process, \textit{``các dịch xã hội cơ bản như thông tin'' (basic social services such as information)}, while our PhraseTransformer can recognize it.  Besides, in the second row, the output of the original Transformer is lack the smoothness of the natural language \textit{\textit{``đặt mức độ suy giảm khủng hoảng trung bình''}} that makes the translated sentence is  misunderstood \textit{``4 loại dịch bệnh''}.  We argue that because our PhraseTransformer captured the dependencies between phrases, that make model is better in capture the sub-conditions of the long sentence.  In addition, our PhraseTransformer captured the local context in the phrases which supports improving the meaning representation in the translation process. 
    \begin{CJK*}{UTF8}{gbsn} 
        \begin{table*}[htpb]
            \centering 
            \resizebox{\textwidth}{!}{%
            \begin{tabular}{lp{0.85\textwidth}}
                \toprule 
                %   \multicolumn{1}{c}{}&\textbf{Content} \\\midrule
                %  Train & Dev & Test & Vi & Zh & Vi & Zh\\\midrule  
                Input & 这旨在面向全面保障贫困人口的社会民生权利，提高贫困人口获得信息、就业、医疗、教育、住房、生活用水、卫生、环境等基本社会服务的机会。 \\
                Gold sentence &  Điều này hướng tới đảm bảo toàn diện quyền an sinh xã hội của người nghèo, cải thiện mức độ tiếp cận của người nghèo đối với \textbf{các dịch vụ xã hội cơ bản như thông tin,} việc làm, y tế, giáo dục, nhà ở, nước sinh hoạt và vệ sinh, môi trường.	 \\
               Transformer & Đây là cơ hội để đảm bảo toàn diện các quyền an sinh xã hội của người nghèo, nâng cao việc làm, y tế, nhà ở, nước sinh hoạt, vệ sinh, môi trường.	 \\
               PhraseTransformer & Đây là dịp để đảm bảo toàn diện các quyền an sinh xã hội cho người nghèo, nâng cao \textcolor{red}{các dịch vụ xã hội cơ bản như thông tin,} việc làm, y tế, giáo dục, nhà ở, nước sinh hoạt, vệ sinh, môi trường. \\  \midrule
               Input & 政府将设低危机、中等危机、高危机、最高危机4个疫情级别。	\\
               Gold sentence & Chính phủ đã phân loại \textbf{4 cấp độ dịch:} nguy cơ thấp (bình thường mới), nguy cơ trung bình, nguy cơ cao, nguy cơ rất cao.	 \\ 
               Transformer & Chính phủ sẽ \textcolor{red}{đặt mức độ suy giảm khủng hoảng trung bình,} khủng hoảng cao, nguy cơ cao nhất, \textcolor{red}{4 loại dịch bệnh.} \\
               PhraseTransformer &  Chính phủ \textcolor{red}{sẽ có  4 cấp độ dịch bệnh thấp,} nguy cơ trung bình, nguy cơ cao và khủng hoảng cao nhất.\\
                \bottomrule
                
            \end{tabular}
            }
            \caption{Improvement examples of the PhraseTransformer compared with Transformer in the Chinese to Vietnamese test set.  The red part is the main difference between these models. 
            % The test set is the public test provided at the time of the competition.
            \label{tab:improvement}}  
        \end{table*}  
    \end{CJK*}

\section{Conclusion \label{sec:conclusion}} 
    In this paper, we first applied our PhraseTransformer model to the machine translation task for the pair of languages Vietnamese - Chinese.   In this architecture, the original Transformer Encoder is enhanced by incorporating the phrase dependencies information into the Self-Attention mechanism. Our experimental results showed that our PhaseTransformer beat the original Transformer by a large margin in both translation directions. In future work, we would like to apply our model architecture to other NLP tasks and explore other effective phrase modeling methods to achieve better results.
    
    % annonymous \paragraph{\textbf{Acknowledgments.}} We would like to thank XUE Jieying at the Japan Advanced Institute of Science and Technology (JAIST) for supporting us in this paper representation. 
    
\bibliography{ref}
\bibliographystyle{acl_natbib}
 
\end{document}